\documentclass[11pt,a4paper]{article}
\usepackage[hyperref]{acl2021}
\usepackage{times}
\usepackage{latexsym}

\usepackage{microtype}

\aclfinalcopy 

\usepackage{subcaption}

\usepackage{multirow}
\usepackage{amsmath}
\usepackage{capt-of}
\usepackage{tabularx}
\usepackage{epsfig}
\usepackage{amssymb}
\usepackage{amsfonts}
\usepackage{booktabs}
\usepackage{scalerel}
\usepackage[inline]{enumitem}
\usepackage{listings}
\usepackage{varwidth}
\usepackage[export]{adjustbox}
\usepackage{tikz}
\usetikzlibrary{tikzmark}
\usepackage{cleveref}

\usepackage{stmaryrd}
\usepackage{bbm}

\usepackage{algorithm}
\usepackage[noend]{algpseudocode}

\definecolor{deepblue}{rgb}{0,0,0.5}
\definecolor{officeblue}{RGB}{0,102,204}
\definecolor{deepred}{rgb}{0.6,0,0}
\definecolor{deepgreen}{rgb}{0,0.5,0}
\definecolor{mybrickred}{RGB}{182,50,28}

\definecolor{fillcolor}{RGB}{216,217,252}

\algnewcommand\algorithmicrequireb{{\hspace{0.85cm}}}
\algnewcommand\INPTDESCB{\item[\algorithmicrequireb]}

\algnewcommand\algorithmicfuncdesc{\textbf{Function:}}
\algnewcommand\FUNCDESC{\item[\algorithmicfuncdesc]}
\algnewcommand\algorithmicfuncdescb{{\hspace{1.48cm}}}
\algnewcommand\FUNCDESCB{\item[\algorithmicfuncdescb]}
\algnewcommand{\algorithmicgoto}{\textbf{goto}}
\algnewcommand{\Goto}[1]{\algorithmicgoto~\ref{#1}}

\usepackage{amsmath,amsfonts,bm}

\def\eqref#1{equation~\ref{#1}}

\def\1{\bm{1}}
\newcommand{\train}{\mathcal{D}}

\def\vtheta{{\bm{\theta}}}
\def\va{{\bm{a}}}

\def\vg{{\bm{g}}}
\def\vh{{\bm{h}}}

\def\vq{{\bm{q}}}

\def\vu{{\bm{u}}}
\def\vv{{\bm{v}}}

\DeclareMathAlphabet{\mathsfit}{\encodingdefault}{\sfdefault}{m}{sl}
\SetMathAlphabet{\mathsfit}{bold}{\encodingdefault}{\sfdefault}{bx}{n}

\newcommand{\Ls}{\mathcal{L}}

\newcommand{\softmax}{\mathrm{softmax}}

\newcommand{\diag}{\mathrm{diag}}

\newcommand\our{\textsc{XLM-Align}}

\newcommand*\samethanks[1][\value{footnote}]{\footnotemark[#1]}

\title{Improving Pretrained Cross-Lingual Language Models via \\ Self-Labeled Word Alignment}

\author{Zewen Chi$^{\dag}$\thanks{\ \  Contribution during internship at Microsoft Research.}\textbf{,}~~Li Dong$^\ddag$\textbf{,}~~Bo Zheng$^\ddag$\samethanks[1]\textbf{,}~~Shaohan Huang$^\ddag$\\
~~\textbf{Xian-Ling Mao}$^\dag$\textbf{,}~~\textbf{Heyan Huang}$^\dag$\textbf{,}~~\textbf{Furu Wei}$^\ddag$\\
$^\dag$Beijing Institute of Technology \\
$^\ddag$Microsoft Research \\
\texttt{\{czw,maoxl,hhy63\}@bit.edu.cn}
\\\texttt{\{lidong1,v-zhebo,shaohanh,fuwei\}@microsoft.com} \\}

\date{}

\begin{document}
\maketitle
\begin{abstract}
The cross-lingual language models are typically pretrained with masked language modeling on multilingual text or parallel sentences. In this paper, we introduce denoising word alignment as a new cross-lingual pre-training task. Specifically, the model first self-labels word alignments for parallel sentences. Then we randomly mask tokens in a bitext pair. Given a masked token, the model uses a pointer network to predict the aligned token in the other language. We alternately perform the above two steps in an expectation-maximization manner. Experimental results show that our method improves cross-lingual transferability on various datasets, especially on the token-level tasks, such as question answering, and structured prediction. Moreover, the model can serve as a pretrained word aligner, which achieves reasonably low error rates on the alignment benchmarks. The code and pretrained parameters are available at \url{github.com/CZWin32768/XLM-Align}.
\end{abstract}

\section{Introduction}

Despite the current advances in NLP, most applications and resources are still English-centric, making non-English users hard to access.
Therefore, it is essential to build cross-lingual transferable models that can learn from the training data in high-resource languages and generalize on low-resource languages.
Recently, pretrained cross-lingual language models have shown their effectiveness for cross-lingual transfer. By pre-training on monolingual text and parallel sentences, the models provide significant improvements on a wide range of cross-lingual end tasks~\cite{xlm,xlmr,mbart,xnlg,infoxlm,xlme}.

Cross-lingual language model pre-training is typically achieved by learning various pretext tasks on monolingual and parallel corpora.
By simply learning masked language modeling (MLM; \citealt{bert}) on monolingual text of multiple languages, the models surprisingly achieve competitive results on cross-lingual tasks~\cite{wu2019beto,xlingual:mbert:iclr20}.
Besides, several pretext tasks are proposed to utilize parallel corpora to learn better sentence-level cross-lingual representations~\cite{xlm,infoxlm,hu2020explicit}.
For example, the translation language modeling (TLM; \citealt{xlm}) task performs MLM on the concatenated parallel sentences, which implicitly enhances cross-lingual transferability.
However, most pretext tasks either learn alignment at the sentence level or implicitly encourage cross-lingual alignment, leaving explicit fine-grained alignment task not fully explored.

In this paper, we introduce a new cross-lingual pre-training task, named as \textit{denoising word alignment}.
Rather than relying on external word aligners trained on parallel corpora~\cite{Cao2020Multilingual,zhao2020inducing,wu2020explicit}, we utilize self-labeled alignments in our task.
During pre-training, we alternately self-label word alignments and conduct the denoising word alignment task in an expectation-maximization manner.
Specifically, the model first self-labels word alignments for a translation pair.
Then we randomly mask tokens in the bitext sentence, which is used as the perturbed input for denosing word alignment.
For each masked token, the model learns a pointer network to predict the self-labeled alignments in the other language.
We repeat the above two steps to iteratively boost the bitext alignment knowledge for cross-lingual pre-training.

We conduct extensive experiments on a wide range of cross-lingual understanding tasks.
Experimental results show that our model outperforms the baseline models on various datasets, particularly on the token-level tasks such as question answering and structured prediction.
Moreover, our model can also serve as a multilingual word aligner, which achieves reasonable low error rates on the bitext alignment benchmarks.

Our contributions are summarized as follows:
\begin{itemize}
\item We present a cross-lingual pre-training paradigm that alternately self-labels and predicts word alignments.
\item We introduce a pre-training task, denoising word alignment, which predicts word alignments from perturbed translation pairs.
\item We propose a word alignment algorithm that formulates the word alignment problem as optimal transport.
\item We demonstrate that our explicit alignment objective is effective for cross-lingual transfer.
\end{itemize}

\section{Related Work}

\paragraph{Cross-lingual LM pre-training}
Pretrained with masked language modeling (MLM;~\citealt{bert}) on monolingual text, multilingual BERT (mBERT;~\citealt{bert}) and XLM-R~\cite{xlmr} produce promising results on cross-lingual transfer benchmarks~\cite{xtreme}. mT5~\cite{mt5} learns a multilingual version of T5~\cite{t5} with text-to-text tasks. In addition to monolingual text, several methods utilize parallel corpora to improve cross-lingual transferability. XLM~\cite{xlm} presents the translation language modeling (TLM) task that performs MLM on concatenated translation pairs. ALM~\cite{alm} introduces code-switched sequences into cross-lingual LM pre-training. Unicoder~\cite{unicoder} employs three cross-lingual tasks to learn mappings among languages. From an information-theoretic perspective, InfoXLM~\cite{infoxlm} proposes the cross-lingual contrastive learning task to align sentence-level representations. Additionally, AMBER~\cite{hu2020explicit} introduces an alignment objective that minimizes the distance between the forward and backward attention matrices. 
More recently, Ernie-M~\cite{erniem} presents the back-translation masked language modeling task that generates pseudo parallel sentence pairs for learning TLM, which provides better utilization of monolingual corpus. 
VECO~\cite{veco} pretrains a unified cross-lingual language model for both NLU and NLG.
mT6~\cite{mt6} improves the multilingual text-to-text transformer with translation pairs.

Notably, word-aligned BERT models~\cite{Cao2020Multilingual,zhao2020inducing} finetune mBERT by an explicit alignment objective that minimizes the distance between aligned tokens. \citet{wu2020explicit} exploit contrastive learning to improve the explicit alignment objectives. However, \citet{wu2020explicit} show that these explicit alignment objectives do not improve cross-lingual representations under a more extensive evaluation. Moreover, these models are restricted to stay close to their original pretrained values, which is not applicable for large-scale pre-training.
On the contrary, we demonstrate that employing our explicit alignment objective in large-scale pre-training can provide consistent improvements over baseline models.

\paragraph{Word alignment}
The IBM models~\cite{ibmmodel} are statistical models for modeling the translation process that can extract word alignments between sentence pairs. A large number of word alignment models are based on the IBM models~\cite{och2003systematic,mermer2011bayesian,fastalign,ostling2016efficient}.
Recent studies have shown that word alignments can be extracted from neural machine translation models~\cite{ghader-monz-2017-attention,koehn-knowles-2017-six,li2019word} or from pretrained cross-lingual LMs~\cite{simalign,nagata2020}.

\begin{figure*}[t]
\centering
\includegraphics[width=0.94\textwidth]{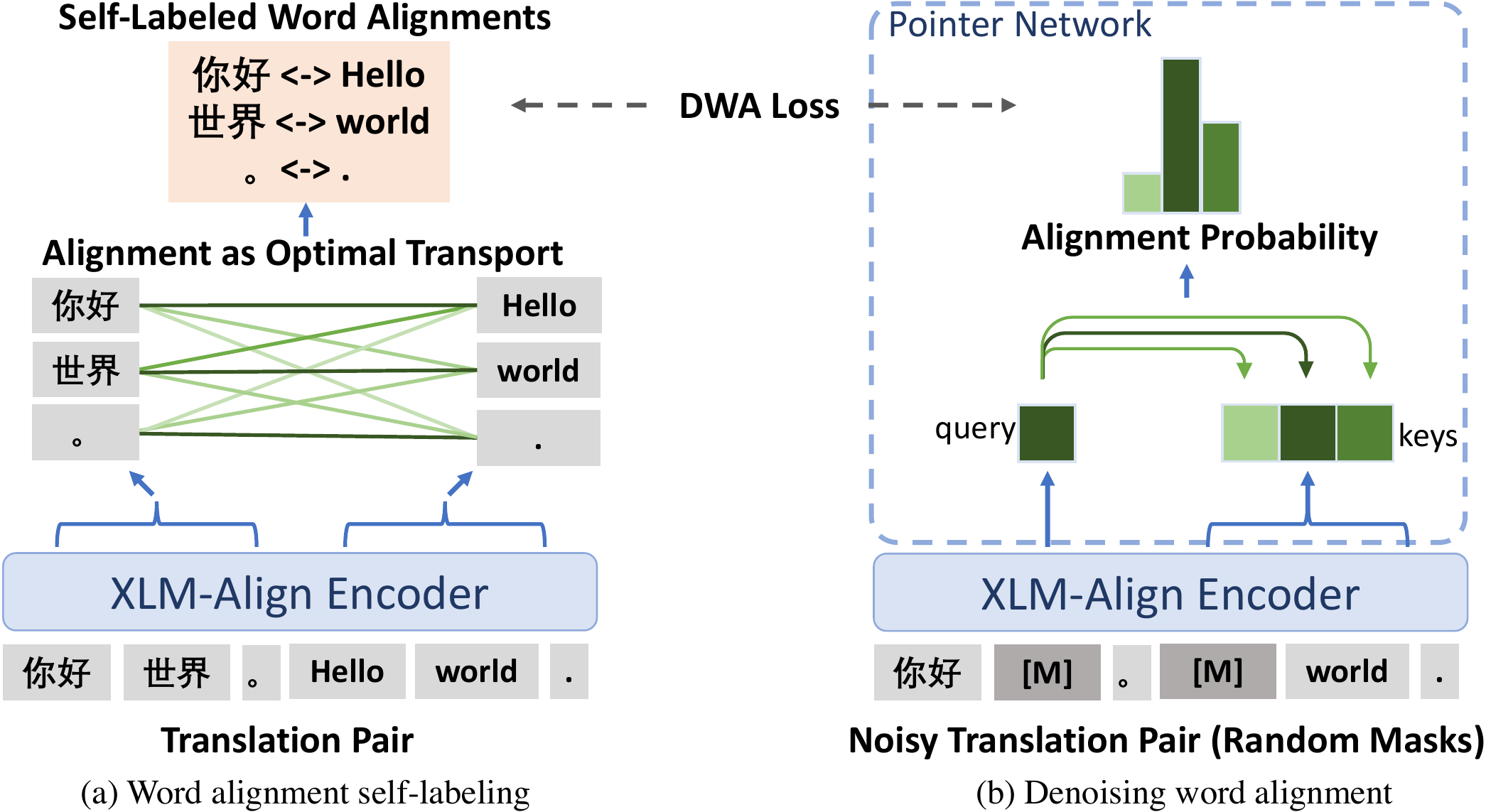}
\caption{An overview of our method. \our{} is pretrained in an expectation-maximization manner with two alternating steps. (a) Word alignment self-labeling: we formulate word alignment as an optimal transport problem, and self-labels word alignments of the input translation pair on-the-fly; (b) Denoising word alignment: we update the model parameters with the denoising word alignment task, where the model uses a pointer network to predict the aligned tokens from the perturbed translation pair.}
\label{fig:method}
\end{figure*}

\section{Method}

Figure~\ref{fig:method} illustrates an overview of our method for pre-training our cross-lingual LM, which is called \our{}. 
\our{} is pretrained in an expectation-maximization manner with two alternating steps, which are word alignment self-labeling and denoising word alignment. 
We first formulate word alignment as an optimal transport problem, and self-label word alignments of the input translation pair on-the-fly.
Then, we update the model parameters with the denoising word alignment task, where the model uses a pointer network~\cite{pointer} to predict the aligned tokens from the perturbed translation pair.

\subsection{Word Alignment Self-Labeling}
\label{sec:e-step}
The goal of word alignment self-labeling is to estimate the word alignments of the input translation pair on-the-fly, given the current \our{} model.
Given a source sentence $\mathcal{S} = s_1 \dots  s_i  \dots  s_n $ and a target sentence $\mathcal{T} = t_1  \dots  t_j  \dots  t_m $, we model the word alignment between $\mathcal{S}$ and $\mathcal{T}$ as a doubly stochastic matrix $A \in \mathbb{R}^{n \times m}_+$ such that the rows and the columns all sum to $1$, where $A_{ij}$ stands for the probability of the alignment between $s_i$ and $t_j$.
The rows and the columns of $A$ represent probability distributions of the forward alignment and the backward alignment, respectively.
To measure the similarity between two tokens from $\mathcal{S}$ and $\mathcal{T}$, we define a metric function $f_\text{sim}$ by using cross-lingual representations produced by \our{}:
\begin{align}
f_\text{sim}(s_i, t_j) = \log \max ( \epsilon, \vh_i^\top\vh_j )
\end{align}
where $\epsilon$ is a constant to avoid negative values in the $\log$ function, and $\vh_i$ is the hidden vector of the $i$-th token by encoding the concatenated sequence of $\mathcal{S}$ and $\mathcal{T}$ with \our{}. Empirically, the metric function produces a high similarity score if the two input tokens are semantically similar.

The word alignment problem is formulated as finding $A$ that maximizes the sentence similarity between $\mathcal{S}$ and $\mathcal{T}$:
\begin{align}
\label{eq:wa}
\max_{A} \sum_{i=1}^{n} \sum_{j=1}^{m} A_{ij} f_\text{sim}(s_i, t_j)
\end{align}
We can find that Eq. (\ref{eq:wa}) is identical to the regularized optimal transport problem \cite{ot}, if we add an entropic regularization to $A$:
\begin{align}
\label{eq:ot}
\max_{A} \sum_{i=1}^{n} \sum_{j=1}^{m} A_{ij} f_\text{sim}(s_i, t_j) - \mu A_{ij}\log A_{ij}
\end{align}
Eq. (\ref{eq:ot}) has a unique solution $A^*$ such that
\begin{align}
    A^* &={\diag}(\vu) K {\diag}(\vv) \\
    K_{ij} &=e^{f_\text{sim}(s_i,t_j)/\mu}
\end{align}
where $\vu \in \mathbb{R}_+^{n}, \vv \in \mathbb{R}_+^{m}, K \in \mathbb{R}_+^{n \times m}$. According to Sinkhorn's algorithm \cite{ot}, the variables $\vu$ and $\vv$ can be calculated by the following iterations:
\begin{align}
    \vu^{t+1}=\frac{\1_n}{K\vv^t},~~~\vv^{t+1}=\frac{\1_m}{K^\top\vu^{t+1}}
\end{align}
where $\vv^{t}$ can be initialized by $\vv^{t=0}=\1_m$.

With the solved stochastic matrix $A^*$, we can produce the forward word alignments $\overrightarrow{\mathcal{A}}$ by applying argmax over rows:
\begin{align}
\label{eq:argmax}
    \overrightarrow{\mathcal{A}} = \{ (i,j) ~| ~j = \arg\max_{k} A^*_{ik} \}
\end{align}
Similarly, the backward word alignments $\overleftarrow{\mathcal{A}}$ can be computed by applying argmax over columns.
To obtain high-precision alignment labels, we adopt an iterative alignment filtering operation. We initialize the alignment labels $\mathcal{A}$ as $\emptyset$. In each iteration, we follow the procedure of Itermax~\cite{simalign} that first computes $\overrightarrow{\mathcal{A}}$ and $\overleftarrow{\mathcal{A}}$ by Eq. (\ref{eq:argmax}).
Then, the alignment labels are updated by:
\begin{align}
    \mathcal{A} \gets \mathcal{A} \cup (\overrightarrow{\mathcal{A}} \cap \overleftarrow{\mathcal{A}})
\label{eq:agree}
\end{align}
Finally, $A^*$ is updated by:
\begin{align}
A^*_{ij} \gets
\begin{cases}
0, & (i,j) \in \mathcal{A} \\
\alpha A^*_{ij}, & \exists k~~ (i, k) \in \mathcal{A} \lor (k, j) \in \mathcal{A} \\
A^*_{ij}, & \text{others}
\end{cases}
\end{align}
where $\alpha$ is a discount factor. 
After several iterations, we obtain the final self-labeled word alignments $\mathcal{A}$.

\subsection{Denoising Word Alignment}

After self-labeling word alignments, we update the model parameters with the denoising word alignment (DWA) task. The goal of DWA is to predict the word alignments from the perturbed version of the input translation pair.

Consider the perturbed version of the input translation pair $(\mathcal{S}^*, \mathcal{T}^*)$ constructed by randomly replacing the tokens with masks.
We first encode the translation pair into hidden vectors $\vh^*$ with the \our{} encoder:
\begin{align}
    \vh^*_1 \dots \vh^*_{n+m} = \text{encoder}([\mathcal{S}^*, \mathcal{T}^*])
\end{align}
where $[\mathcal{S}^*, \mathcal{T}^*]$ is the concatenated sequence of $\mathcal{S}^*$ and $\mathcal{T}^*$ with the length of $n + m$.
Then, we build a pointer network upon the \our{} encoder that predicts the word alignments. Specifically, for the $i$-th source token, we use $\vh^*_i$ as the query vector and $\vh^*_{n+1}$, $\dots$, $\vh^*_{n+m}$ as the key vectors. Given the query and key vectors, the forward alignment probability $\va_i$ is computed by the scaled dot-product attention~\cite{transformer}:
\begin{align}
    \va_i &= {\softmax}(\frac{\vq_i^\top K}{\sqrt{d_h}}) \\
    \vq_i &= \mathrm{linear} ( \vh^*_i ) \\
    K &= \mathrm{linear} ( [\vh^{*}_{n+1} \dots \vh^{*}_{n+m}] )
\end{align}
where $d_h$ is the dimension of the hidden vectors. Similarly, the backward alignment probability can be computed by above equations if we use target tokens as the query vectors and $\vh^*_{1} \dots \vh^*_{n}$ as key vectors. 
Notice that we only consider the self-labeled and masked positions as queries. Formally, we use the following query positions in the pointer network:
\begin{align}
\mathcal{P} = \left \{ i | (i, \cdot) \in \mathcal{A} ~\lor~ (\cdot, i) \in \mathcal{A} \right \} \cap \mathcal{M}
\end{align}
where $\mathcal{M}$ is the set of masked positions.
The training objective is to minimize the cross-entropy between the alignment probabilities and the self-labeled word alignments:
\begin{align}
    \Ls_\text{DWA} = \sum_{i \in \mathcal{P}} \text{CE}(\va_i, \mathcal{A}(i))
\end{align}
where $\text{CE}(\cdot,\cdot)$ stands for the cross-entropy loss, and $\mathcal{A}(i)$ is the self-labeled aligned position of the $i$-th token. 

\begin{algorithm}
\caption{Pre-training \our{}}
\label{alg:em}
\begin{algorithmic}[1]
\Require Multilingual corpus $\train_\text{m}$, parallel corpus $\train_\text{p}$, learning rate $\tau$
\Ensure \our{} parameters $\vtheta$
\State Initialize $\vtheta$ with cold-start pre-training
\While{not converged}
\State $\mathcal{X} \sim \train_\text{m}, ~~(\mathcal{S}, \mathcal{T}) \sim \train_\text{p}$
\State $\mathcal{A} \gets f_\text{self-labeling}(\mathcal{S}, \mathcal{T}; \vtheta)$
\State $\vg \gets \nabla_\vtheta \Ls_\text{MLM}(\mathcal{X}) + \nabla_\vtheta \Ls_\text{TLM}(\mathcal{S}, \mathcal{T}) + \nabla_\vtheta \Ls_\text{DWA}(\mathcal{S}, \mathcal{T}, \mathcal{A})$
\State $\vtheta \gets \vtheta - \tau \vg$
\EndWhile
\end{algorithmic}
\end{algorithm}

\begin{table*}[t]
\centering
\scalebox{0.88}{
\begin{tabular}{lcccccccccccccccccc}
\toprule
\multirow{2}{*}{\bf Model} & \multicolumn{2}{c}{\bf Structured Prediction} & \multicolumn{3}{c}{\bf Question Answering} & \multicolumn{2}{c}{\bf Sentence Classification} & \multirow{2}{*}{\bf Avg} \\
& POS & NER & XQuAD & MLQA & TyDiQA & XNLI & PAWS-X & \\ \midrule
Metrics & F1 & F1 & F1 / EM & F1 / EM & F1 / EM & Acc. & Acc. & \\ 
\midrule
\textsc{mBert}* & 70.3 & 62.2 & 64.5 / 49.4 & 61.4 / 44.2 & 59.7 / 43.9 & 65.4 & 81.9 & 63.1 \\
\textsc{XLM}*  & 70.1 & 61.2 & 59.8 / 44.3 & 48.5 / 32.6 & 43.6 / 29.1 & 69.1 & 80.9 & 58.6 \\
\textsc{mT5}$_\text{base}$ & - & 56.6 & 67.0 / 49.0 & 64.6 / 45.0 & 58.1 / 42.8 & 75.4 & \textbf{87.4} & - \\
\textsc{XLM-R}$_\text{base}$ & 75.6 & 61.8 & 71.9 / 56.4 & 65.1 / 47.2 & 55.4 / 38.3 & 75.0 & 84.9 & 66.4 \\
\our{} & \textbf{76.0} & \textbf{63.7} & \textbf{74.7} / \textbf{59.0} & \textbf{68.1} / \textbf{49.8} & \textbf{62.1} / \textbf{44.8} & \textbf{76.2} & 86.8 & \textbf{68.9} \\
\bottomrule
\end{tabular}}
\caption{Evaluation results on XTREME structured prediction, question answering, and sentence classification tasks. We adopt the cross-lingual transfer setting, where models are only fine-tuned on the English training data but evaluated on all target languages. Results with ``*'' are taken from~\cite{xtreme}. Results of \our{} and XLM-R$_\text{base}$ are averaged over five runs.}
\label{table:overview}
\end{table*}

\subsection{Pre-training \our{}}

We illustrate the pre-training procedure of \our{} in Algorithm~\ref{alg:em}.
In addition to DWA, we also include MLM and TLM for pre-training \our{}, which implicitly encourage the cross-lingual alignment.
The overall loss function is defined as:
\begin{align}
    \Ls_\text{MLM}(\mathcal{X}) + \Ls_\text{TLM}(\mathcal{S}, \mathcal{T}) + \Ls_\text{DWA}(\mathcal{S}, \mathcal{T}, \mathcal{A}) \nonumber
\end{align}
In each iteration, we first sample monolingual text $\mathcal{X}$, and parallel text $(\mathcal{S}, \mathcal{T})$. Then, we self-label word alignments and update the model parameters by learning pretext tasks.
Notice that the model parameters are initialized by a cold-start pre-training to avoid producing low-quality alignment labels. The cold-start pre-training can be accomplished by using a pretrained LM as the model initialization.

\section{Experiments}

\subsection{Pre-training}

Following previous cross-lingual pretrained models~\cite{xlm,xlmr,infoxlm}, we use raw sentences from the Wikipedia dump and CCNet~\cite{ccnet} for MLM, including $94$ languages. For TLM and DWA, we use parallel corpora from MultiUN~\cite{multiun}, IIT Bombay~\cite{iit}, OPUS~\cite{opus}, and WikiMatrix~\cite{wikimatrix}, including $14$ English-centric language pairs. We pretrain a Transformer with $12$ layers and the hidden size of $768$, where the parameters are initialized with XLM-R~\cite{xlmr}. The model is optimized with the Adam optimizer~\cite{adam} for $150$K steps with batch size of $2,048$.
Notice that TLM and DWA share the same forward procedure for encoding the perturbed sentence pair. The pre-training of \our{} takes about six days with two Nvidia DGX-2 stations. More details of the training data and the hyperparameters are in supplementary document.

\subsection{XTREME Benchmark}

XTREME is a multilingual benchmark for evaluating cross-lingual generalization. 
We evaluate our model on $7$ cross-lingual downstream tasks included by XTREME, which can be grouped into $3$ categories:
(1) Structured prediction: part-of-speech tagging on the Universal Dependencies v2.5~\cite{udpos}, and named entity recognition on the WikiAnn~\cite{panx,rahimi2019} dataset;
(2) Question answering: cross-lingual question answering on MLQA~\cite{mlqa} and XQuAD~\cite{xquad}, and gold passage of typologically diverse question answering (TyDiQA-GoldP;~\citealt{tydiqa});
(3) Sentence classification: cross-lingual natural language inference (XNLI;~\citealt{xnli}), and cross-lingual paraphrase adversaries from word scrambling (PAWS-X;~\citealt{pawsx}).

\paragraph{Baselines}
We use the following pretrained cross-lingual LMs as baselines. (1) Multilingual BERT (\textsc{mBert};~\citealt{bert}) is pretrained with masked language modeling (MLM) and next sentence prediction on Wikipedia of 104 languages; (2) XLM~\cite{xlm} is jointly pretrained with MLM on 100 languages and translation language modeling (TLM) on 14 language pairs; (3) \textsc{mT5}~\cite{mt5} is the multilingual version of T5 pretrained with text-to-text tasks; (4) XLM-R~\cite{xlmr} is pretrained with MLM on large-scale CC-100 dataset with long training steps.

\paragraph{Fine-tuning}
Following~\citet{xtreme}, we adopt the zero-shot transfer setting for evaluation, where the models are only fine-tuned on English training data but evaluated on all target languages.
Besides, we only use one model for evaluation on all target languages, rather than selecting different models for each language. The detailed fine-tuning hyperparameters can be found in supplementary document.

\begin{table*}
\centering
\scalebox{0.94}{
\begin{tabular}{lcccccc}
\toprule
\multirow{2}{*}{\bf Alignment Method} & {\bf Pretrained} & \multicolumn{4}{c}{\bf Alignment Error Rate $\downarrow$} & \multirow{2}{*}{\bf Avg} \\
 & \textbf{Model} & en-de & en-fr & en-hi & en-ro &  \\ \midrule
\texttt{fast\_align}~\cite{fastalign} & - & 32.14 & 19.46 & 59.90 & - & - \\
\texttt{SimAlign} - Argmax~\cite{simalign} & XLM-R & 19. & {7.} & 39. & 29. & 24. \\
\texttt{SimAlign} - Itermax~\cite{simalign} & XLM-R & 20. & 9. & 39. & 28. & 24. \\
\texttt{SimAlign} - Itermax (reimplementation) & XLM-R & 20.15 & 10.05 & 38.72 & {27.41} & 24.08 \\
Ours - Optimal Transport (Section~\ref{sec:e-step}) & XLM-R & {17.74} & 7.54 & {37.79} & 27.49 & {22.64} \\ \midrule
\texttt{SimAlign} (reimplementation) & \our{} & 18.93 & 10.33 & \textbf{33.84} & 27.09 & 22.55 \\
Ours - Optimal Transport (Section~\ref{sec:e-step}) & \our{} & \textbf{16.63} & \textbf{6.61} & 33.98 & \textbf{26.97} & \textbf{21.05} \\
\bottomrule
\end{tabular}}
\caption{Evaluation results for word alignment on four English-centric language pairs. We report the alignment error rate scores (lower is better). For both \texttt{SimAlign}~\cite{simalign} and our optimal-transport alignment method, we use the hidden vectors from the $8$-th layer produced by XLM-R$_\text{base}$ or \our{}.  ``(reimplementation)'' is our reimplementation of \texttt{SimAlign}-Itermax.}
\label{table:otalign}
\end{table*}

\paragraph{Results}
In Table~\ref{table:overview}, we present the evaluation results on XTREME structured prediction, question answering, and sentence classification tasks.
It can be observed that our \our{} obtains the best average score over all the baseline models, improving the previous score from 66.4 to 68.9.
It demonstrates that our model learns more transferable representations for the cross-lingual tasks, which is beneficial for building more accessible multilingual NLP applications.
It is worth mentioning that our method brings noticeable improvements on the question answering and the structured prediction tasks.
Compared with XLM-R$_\text{base}$, \our{} provides $6.7\%$ and $1.9\%$ F1 improvements on TyDiQA and NER.
The improvements show that the pretrained \our{} benefits from the explicit word alignment objective, particularly on the structured prediction and question answering tasks that require token-level cross-lingual transfer.
In terms of sentence classification tasks, \our{} also consistently outperforms XLM-R$_\text{base}$.

\begin{figure}
\centering
\includegraphics[width=0.48\textwidth]{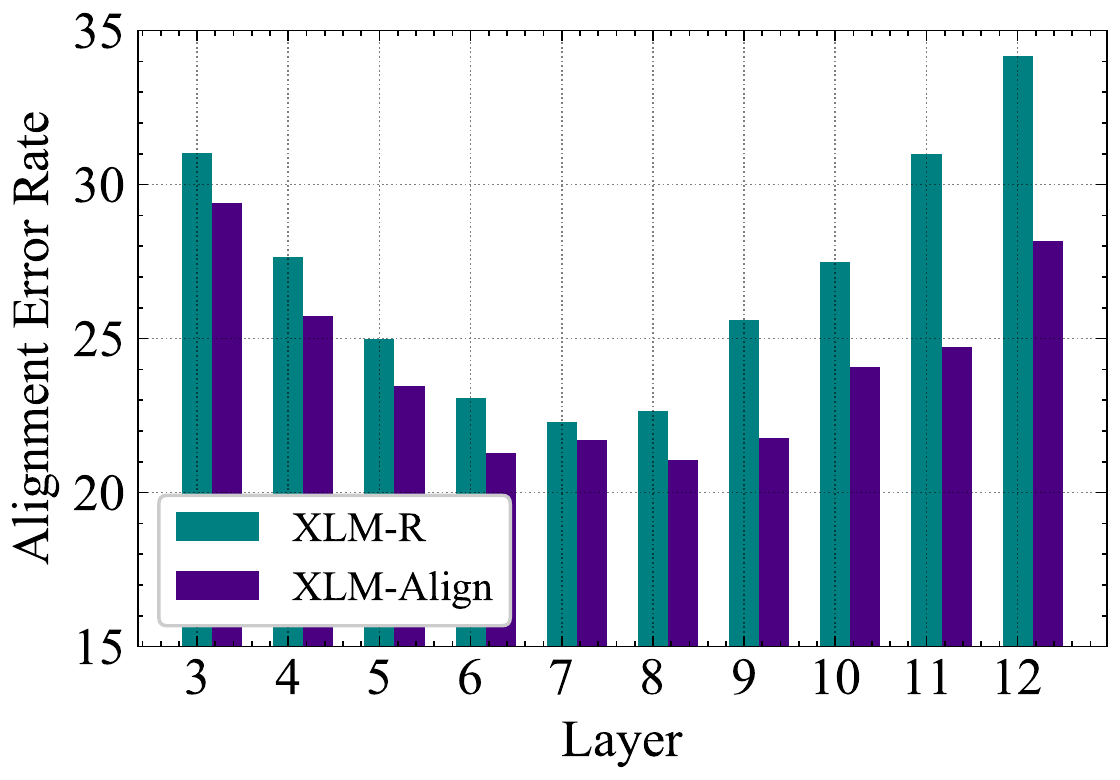}
\label{fig:wa_de-en}
\caption{Evaluation results on word alignment across different layers. We illustrate the averaged AER scores on the test sets of four language pairs. The results of the first two layers are not included due to the high AER.}
\label{fig:wa}
\end{figure}

\subsection{Word Alignment}

Word alignment is the task of finding corresponding word pairs in a parallel sentence.
We conduct evaluations with golden alignments of four language pairs from EuroParl\footnote{\url{www-i6.informatik.rwth-aachen.de/goldAlignment/}}, WPT2003\footnote{\url{web.eecs.umich.edu/~mihalcea/wpt/}}, and WPT2005\footnote{\url{web.eecs.umich.edu/~mihalcea/wpt05/}}, containing 1,244 annotated sentence pairs in total. We use alignment error rate (AER;~\citealt{och2003systematic}) as the evaluation metrics.

\paragraph{Results}
We first explore whether our word alignment self-labeling method is effective for generating high-quality alignment labels. Thus, we compare our method with (1) \texttt{fast\_align}~\cite{fastalign}, a widely-used implementation of IBM Model 2~\cite{och2003systematic}; (2) \texttt{SimAlign}~\cite{simalign}, state-of-the-art unsupervised word alignment method. For a fair comparison, we use the same pretrained LM and hidden layer as in \texttt{SimAlign} to produce sentence representations. In specific, we take the hidden vectors from the $8$-th layer of XLM-R$_\text{base}$ or \our{}, and obtain the alignments following the procedure as described in Section~\ref{sec:e-step}. Since the produced alignments are subword-level, we convert the alignments into word-level by the following rule that ``if two subwords are aligned, the words they belong to are also aligned''.

As shown in Table~\ref{table:otalign}, we report the AER scores on the four language pairs. It can be observed that our optimal-transport method outperforms \texttt{fast\_align} and \texttt{SimAlign}, demonstrating that our method can produce high-quality alignment labels, which is helpful for the DWA task. Moreover, our method consistently outperforms \texttt{SimAlign} when using hidden vectors from both XLM-R$_\text{base}$ and \our{}.

Then, we compare our \our{} with XLM-R$_\text{base}$ on the word alignment task. Empirically, a lower AER indicates that the model learns better cross-lingual representations. From Table~\ref{table:otalign}, \our{} obtains the best AER results over all the four language pairs, reducing the averaged AER from $22.64$ to $21.05$. Besides, under both \texttt{SimAlign} and our optimal-transport method, \our{} provides consistent reduction of AER, demonstrating the effectiveness of our method for learning fine-grained cross-lingual representations.

We also compare \our{} with XLM-R$_\text{base}$ using the hidden vectors from the $3$-th layer to the $12$-th layer.
We illustrate the averaged AER scores in Figure~\ref{fig:wa}. Notice that the results on the first two layers are not presented in the figure because of the high AER.
It can be observed that \our{} consistently improves the results over XLM-R$_\text{base}$ across these layers. 
Moreover, it shows a parabolic trend across the layers of XLM-R$_\text{base}$, which is consistent with the results in~\cite{simalign}. In contrast to XLM-R$_\text{base}$, \our{} alleviates this trend and greatly reduces AER in the last few layers. We believe this property of \our{} brings better cross-lingual transferability on the end tasks.

\section{Analysis}

In this section, we conduct comprehensive ablation studies for a better understanding of our \our{}. To reduce the computational cost, we reduce the batch size to $256$, and pretrain models with $50$K steps in the following experiments.


\begin{table}[t]
\centering
\scalebox{0.84}{
\begin{tabular}{lccccccccccccccccc}
\toprule
\textbf{Models} & \textbf{XNLI} & \textbf{POS} & \textbf{NER} & \textbf{MLQA} & \textbf{Avg} \\ \midrule
XLM-R* & 74.6 & 75.7 & 61.6 & 65.7 & 69.4 \\
\our{} & \textbf{75.2} & 75.6 & \textbf{62.6} & \textbf{66.7} & \textbf{70.0} \\
~~$-$DWA & 75.1 & 75.2 & 62.0 & 65.8 & 69.5  \\
~~$-$TLM & 74.4 & \textbf{76.0} & 60.4 & 66.0 & 69.2 \\ \bottomrule
\end{tabular}}
\caption{Ablation studies on the components of \our{}.
XLM-R* stands for continue-training XLM-R$_\text{base}$ with MLM for fair comparisons.
Results are averaged over five runs.}
\label{table:ablation}
\end{table}

\subsection{Ablation Studies}
We perform ablation studies to understand the components of \our{}, by removing the denoising word alignment loss ($-$DWA), the TLM loss ($-$TLM), or removing both (XLM-R*), which is identical to continue-training XLM-R$_\text{base}$ with MLM.
We evaluate the models on XNLI, POS, NER, and MLQA, and present the results in Table~\ref{table:ablation}.
Comparing $-$TLM with $-$DWA, we find that DWA is more effective for POS and MLQA, while TLM performs better on XNLI and NER. 
Comparing $-$TLM with XLM-R*, it shows that directly learning DWA slightly harms the performance. However, jointly learning DWA with TLM provides remarkable improvements over $-$DWA, especially on the question answering and the structure prediction tasks that requires token-level cross-lingual transfer.
This indicates that TLM potentially improves the quality of self-labeled word alignments, making DWA more effective for cross-lingual transfer.

\begin{table}[t]
\centering
\scalebox{0.94}{
\renewcommand\tabcolsep{5.0pt}
\begin{tabular}{lccccc}
\toprule
\textbf{Layer} & \textbf{XNLI} & \textbf{POS} & \textbf{NER} & \textbf{MLQA} & \textbf{Avg} \\ \midrule
Layer-8 & 75.1 & 75.3 & 61.9 & 66.7 & 69.8 \\
Layer-10 & 75.2 & 75.6 & \textbf{62.6} & 66.7 & 70.0 \\
Layer-12 & \textbf{75.2} & \textbf{75.8} & 62.3 & \textbf{67.0} & \textbf{70.1} \\
\bottomrule
\end{tabular}}
\caption{Results of \our{} with different layers used for word alignment self-labeling during pre-training. Results are averaged over five runs.}
\label{table:walayer}
\end{table}
\subsection{Word Alignment Self-Labeling Layer}
It has been shown that the word alignment performance has a parabolic trend across the layers of mBERT and XLM-R~\cite{simalign}. It indicates that the middle layers produce higher-quality word alignments than the bottom and the top layers.
To explore which layer produces better alignment labels for pre-training, we pretrain three variants of \our{}, where we use the hidden vectors from three different layers for word alignment self-labeling. We use the $8$-th, $10$-th, and $12$-th layers for word alignment self-labeling during the pre-training. We present the evaluation results in Table~\ref{table:walayer}. Surprisingly, although Layer-8 produces higher-quality alignment labels at the beginning of the pre-training, using the alignment labels from the $12$-th layer learns a more transferable \our{} model for cross-lingual end tasks. 

\begin{table}[t]
\centering
\scalebox{0.94}{
\renewcommand\tabcolsep{5.0pt}
\begin{tabular}{lccccc}
\toprule
\textbf{Layer} & \textbf{XNLI} & \textbf{POS} & \textbf{NER} & \textbf{MLQA} & \textbf{Avg} \\ \midrule
Layer-8 &  \textbf{75.4} & 75.3 & 61.7 & 66.2 & 69.7 \\
Layer-10 & 75.1 & 75.6 & \textbf{62.5} & 66.3 & 69.9 \\
Layer-12 & 75.2 & \textbf{75.8} & 62.3 & \textbf{67.0} & \textbf{70.1} \\
\bottomrule
\end{tabular}}
\caption{Results of \our{} with different layers used for denoising word alignment during pre-training. Results are averaged over five runs.}
\label{table:dwalayer}
\end{table}
\subsection{Denoising Word Alignment Layer}
Beyond the self-labeling layer, we also investigate which layer is better for learning the denoising word alignment task. Recent studies have shown that it is beneficial to learn sentence-level cross-lingual alignment at a middle layer~\cite{infoxlm}. Therefore, we pretrain \our{} models by using three different layers for DWA, that is, using the hidden vectors of middle layers as the input of the pointer network. We compare the evaluation results of the three models in Table~\ref{table:dwalayer}. It can be found that learning DWA at Layer-$8$ improves XNLI while learning DWA at higher layers produces better performance on the other three tasks. It suggests that, compared with sentence-level pretext tasks that prefers middle layers, the DWA task should be applied at top layers.

\begin{table}[t]
\centering
\scalebox{0.92}{
\begin{tabular}{lccccc}
\toprule
\textbf{Filtering} & \textbf{XNLI} & \textbf{POS} & \textbf{NER} & \textbf{MLQA} & \textbf{Avg} \\ \midrule
Enable  & \textbf{75.2} & \textbf{75.6} & \textbf{62.6} & \textbf{66.7} & \textbf{70.0} \\
Disable & 74.2 & 75.3 & 61.6 & 65.3 & 69.1 \\
\bottomrule
\end{tabular}}
\caption{Effects of alignment filtering in word alignment self-labeling. 
Results are averaged over five runs.}
\label{table:vs}
\end{table}

\subsection{Effects of Alignment Filtering}
Although our self-labeling method produces high-quality alignment labels, the alignment filtering operation can potentially make some of the tokens unaligned, which reduces the example efficiency.
Thus, we explore whether the alignment filtering is beneficial for pre-training \our{}.
To this end, we pretrain an \our{} model without alignment filtering. In specific, we use the union set of the forward and backward alignments as the self-labeled alignments so that all tokens are aligned at least once.
The forward and backward alignments are obtained by applying the argmax function over rows and columns of $A^*$, respectively. Empirically, the alignment filtering operation generates high-precision yet fewer labels, while removing the filtering promises more labels but introduces low-confident labels. 
In Table~\ref{table:vs}, we compare the results of the models with or without alignment filtering.
It can be observed that the alignment filtering operation improves the performance on the end tasks.
This demonstrates that it is necessary to use high-precision labels for learning the denoising word alignment task.
On the contrary, using perturbed alignment labels in pre-training harms the performance on the end tasks.

\begin{table}[t]
\centering
\scalebox{0.9}{
\renewcommand\tabcolsep{5.0pt}
\begin{tabular}{lccccc}
\toprule
\textbf{Position} & \textbf{XNLI} & \textbf{POS} & \textbf{NER} & \textbf{MLQA} & \textbf{Avg} \\ \midrule
masked & 75.2 & 75.6 & \textbf{62.6} & \textbf{66.7} & \textbf{70.0} \\
unmasked & \textbf{75.5} & 75.5 & 62.0 & 66.5 & 69.8 \\
all-aligned & 75.3 & \textbf{75.9} & 61.6 & \textbf{66.7} & 69.9 \\
no-query & 75.1 & 75.2 & 62.0 & 65.8 & 69.5 \\
\bottomrule
\end{tabular}}
\caption{Effects of the query positions in the pointer network for denoising word alignment. Results are averaged over five runs.}
\label{table:tqamask}
\end{table}

\subsection{Effects of DWA Query Positions}
In the denoising word alignment task, we always use the hidden vectors of the masked positions as the query vectors in the pointer network.
To explore the impact of the DWA query positions, we compare three different query positions in Table~\ref{table:tqamask}: (1) \textit{masked}: only using the masked tokens as queries; (2) \textit{unmasked}: randomly using $15$\% of the unmasked tokens as queries; (3) \textit{all-aligned}: for each self-labeled aligned pair, randomly using one of the two tokens as a query. Also, we include the \textit{no-query} baseline that does not use any queries, which is identical to removing DWA. It can be observed that using all the three query positions improves the performance over the \textit{no-query} baseline.
Moreover, using the masked positions as queries achieves better results than the other two positions, demonstrating the effectiveness of the masked query positions.

\section{Discussion}

In this paper, we introduce denoising word alignment as a new cross-lingual pre-training task. By alternately self-labeling and predicting word alignments, our \our{} model learns transferable cross-lingual representations.
Experimental results show that our method improves the cross-lingual transferability on a wide range of tasks, particularly on the token-level tasks such as question answering and structured prediction.

Despite the effectiveness for learning cross-lingual transferable representations, our method also has the limitation that requires a cold-start pre-training to prevent the model from producing low-quality alignment labels.
In our experiments, we also try to pretrain \our{} from scratch, i.e., without cold-start pre-training. However, the DWA task does not work very well due to the low-quality of self-labeled alignments. Thus, we recommend continue-training \our{} on the basis of other pretrained cross-lingual language models. For future work, we would like to research on removing this restriction so that the model can learn 
word alignments from scratch.

\bibliographystyle{acl_natbib}
\bibliography{cxlm}

\appendix

\section{Pre-Training Data}

We use raw sentences from the Wikipedia dump and CCNet\footnote{\url{https://github.com/facebookresearch/cc_net}} as monolingual corpora. The CCNet corpus we use is reconstructed following ~\cite{xlmr} to reproduce the CC-100 corpus. 
The resulting corpus contains $94$ languages.
Table~\ref{table:cc} and Table~\ref{table:wiki} report the language codes and data size of CCNet and Wikipedia dump.
Notice that several languages share the same ISO language codes, e.g., zh represents both Simplified Chinese and Traditional Chinese.
Besides, Table~\ref{table:parallel:data} shows the statistics of our parallel corpora.

\begin{table}[ht]
\centering
\scriptsize
\begin{tabular}{crcrcr}
\toprule
Code & Size (GB) & Code & Size (GB) & Code & Size (GB) \\ \cmidrule(r){1-2}\cmidrule{3-4}\cmidrule(l){5-6}
af & 0.2 & hr & 1.4 & pa & 0.8 \\
am & 0.4 & hu & 9.5 & pl & 28.6 \\
ar & 16.1 & hy & 0.7 & ps & 0.4 \\
as & 0.1 & id & 17.2 & pt & 39.4 \\
az & 0.8 & is & 0.5 & ro & 11.0 \\
ba & 0.2 & it & 47.2 & ru & 253.3 \\
be & 0.5 & ja & 86.8 & sa & 0.2 \\
bg & 7.0 & ka & 1.0 & sd & 0.2 \\
bn & 5.5 & kk & 0.6 & si & 1.3 \\
ca & 3.0 & km & 0.2 & sk & 13.6 \\
ckb & 0.6 & kn & 0.3 & sl & 6.2 \\
cs & 14.9 & ko & 40.0 & sq & 3.0 \\
cy & 0.4 & ky & 0.5 & sr & 7.2 \\
da & 6.9 & la & 0.3 & sv & 60.4 \\
de & 99.0 & lo & 0.2 & sw & 0.3 \\
el & 13.1 & lt & 2.3 & ta & 7.9 \\
en & 731.6 & lv & 1.3 & te & 2.3 \\
eo & 0.5 & mk & 0.6 & tg & 0.7 \\
es & 85.6 & ml & 1.3 & th & 33.0 \\
et & 1.4 & mn & 0.4 & tl & 1.2 \\
eu & 1.0 & mr & 0.5 & tr & 56.4 \\
fa & 19.0 & ms & 0.7 & tt & 0.6 \\
fi & 5.9 & mt & 0.2 & ug & 0.2 \\
fr & 89.9 & my & 0.4 & uk & 13.4 \\
ga & 0.2 & ne & 0.6 & ur & 3.0 \\
gl & 1.5 & nl & 25.9 & uz & 0.1 \\
gu & 0.3 & nn & 0.4 & vi & 74.5 \\
he & 4.4 & no & 5.5 & yi & 0.3 \\
hi & 5.0 & or & 0.3 & zh & 96.8 \\
\bottomrule
\end{tabular}
\caption{The statistics of CCNet used for pre-training.}
\label{table:cc}
\end{table}

\begin{table}[t]
\centering
\scriptsize
\begin{tabular}{crcrcr}
\toprule
Code & Size (GB) & Code & Size (GB) & Code & Size (GB) \\ \cmidrule(r){1-2}\cmidrule{3-4}\cmidrule(l){5-6}
af & 0.12 & hr & 0.28 & pa & 0.10 \\
am & 0.01 & hu & 0.80 & pl & 1.55 \\
ar & 1.29 & hy & 0.60 & ps & 0.04 \\
as & 0.04 & id & 0.52 & pt & 1.50 \\
az & 0.24 & is & 0.05 & ro & 0.42 \\
ba & 0.13 & it & 2.70 & ru & 5.63 \\
be & 0.31 & ja & 2.65 & sa & 0.04 \\
bg & 0.62 & ka & 0.37 & sd & 0.02 \\
bn & 0.41 & kk & 0.29 & si & 0.09 \\
ca & 1.10 & km & 0.12 & sk & 0.21 \\
ckb & 0.00 & kn & 0.25 & sl & 0.21 \\
cs & 0.81 & ko & 0.56 & sq & 0.11 \\
cy & 0.06 & ky & 0.10 & sr & 0.74 \\
da & 0.33 & la & 0.05 & sv & 1.70 \\
de & 5.43 & lo & 0.01 & sw & 0.03 \\
el & 0.73 & lt & 0.19 & ta & 0.46 \\
en & 12.58 & lv & 0.12 & te & 0.45 \\
eo & 0.25 & mk & 0.34 & tg & 0.04 \\
es & 3.38 & ml & 0.28 & th & 0.52 \\
et & 0.23 & mn & 0.05 & tl & 0.04 \\
eu & 0.24 & mr & 0.10 & tr & 0.43 \\
fa & 0.66 & ms & 0.20 & tt & 0.09 \\
fi & 0.68 & mt & 0.01 & ug & 0.03 \\
fr & 4.00 & my & 0.15 & uk & 2.43 \\
ga & 0.03 & ne & 0.06 & ur & 0.13 \\
gl & 0.27 & nl & 1.38 & uz & 0.06 \\
gu & 0.09 & nn & 0.13 & vi & 0.76 \\
he & 1.11 & no & 0.54 & yi & 0.02 \\
hi & 0.38 & or & 0.04 & zh & 1.08 \\
\bottomrule
\end{tabular}
\caption{The statistics of Wikipedia dump used for pre-training.}
\label{table:wiki}
\end{table}

\begin{table}[t]
\centering
\scriptsize
\begin{tabular}{crcr}
\toprule
ISO Code & Size (GB) & ISO Code & Size (GB) \\ \midrule
en-ar & 5.88 & en-ru & 7.72 \\
en-bg & 0.49 & en-sw & 0.06 \\
en-de & 4.21 & en-th & 0.47 \\
en-el & 2.28 & en-tr & 0.34 \\
en-es & 7.09 & en-ur & 0.39 \\
en-fr & 7.63 & en-vi & 0.86 \\
en-hi & 0.62 & en-zh & 4.02 \\
\bottomrule
\end{tabular}
\caption{Parallel data used for pre-training.}
\label{table:parallel:data}
\end{table}

\section{Hyperparameters for Pre-Training}

As shown in Table~\ref{table:pt-hparam}, we present the hyperparameters for pre-training \our{}. We use the same vocabulary with XLM-R \cite{xlmr}.

\begin{table}[t]
\centering
\small
\renewcommand\tabcolsep{2.8pt}
\begin{tabular}{lr}
\toprule
Hyperparameters & Value \\ \midrule
Layers & 12 \\
Hidden size & 768 \\
FFN inner hidden size & 3,072 \\
Attention heads & 12 \\
Training steps & 150K \\
Batch size & 2,048 \\
Adam $\epsilon$ & 1e-6 \\
Adam $\beta$ & (0.9, 0.98) \\
Learning rate & 2e-4 \\
Learning rate schedule & Linear \\
Warmup steps & 10,000 \\
Gradient clipping & 1.0 \\
Weight decay & 0.01 \\
Self-labeling layer & 10 \\
Entropic regularization $\mu$ & 1.0 \\
Sinkhorn iterations & 2 \\
Alignment filtering iterations & 2 \\
Alignment filtering $\alpha$ & 0.9 \\
\bottomrule
\end{tabular}
\caption{Hyperparameters used for pre-training \our{}.}
\label{table:pt-hparam}
\end{table}

\section{Hyperparameters for Fine-Tuning}

In Table~\ref{table:hparam}, we present the hyperparameters for fine-tuning XLM-R$_\text{base}$ and \our{} on the XTREME end tasks. For each task, the hyperparameters are searched on the joint validation set of all languages. 

\begin{table*}
\centering
\small
\begin{tabular}{lrrrrrrr}
\toprule
& POS & NER & XQuAD & MLQA & TyDiQA & XNLI & PAWS-X \\ \midrule
Batch size & \{8,16,32\} & 8 & 32 & 32 & 32 & 32 & 32 \\
Learning rate & \{1,2,3\}e-5 & \{5,...,9\}e-6 & \{2,3,4\}e-5 & \{2,3,4\}e-5 & \{2,3\}e-5 & \{5,...,8\}e-6 & \{1,2\}e-5 \\
LR schedule & Linear & Linear & Linear & Linear & Linear & Linear & Linear \\
Warmup & 10\% & 10\% & 10\% & 10\% & 10\% & 12,500 steps & 10\% \\
Weight decay & 0 & 0 & 0 & 0 & 0 & 0 & 0 \\
Epochs & 10 & 10 & 4 & \{2,3,4\} & \{5,10,15,20\} & 10 & 10\\
\bottomrule
\end{tabular}
\caption{Hyperparameters used for fine-tuning XLM-R$_\text{base}$ and \our{} on the XTREME end tasks.}
\label{table:hparam}
\end{table*}

\section{Detailed Results on XTREME}

We present the detailed results of \our{} on XTREME in Table~\ref{table:udpos}-\ref{table:pawsx}.

\begin{table*}[b]
\centering
\small
\renewcommand\tabcolsep{4.0pt}
\scalebox{0.9}{
\begin{tabular}{lccccccccccccccccc}
\toprule
 Model & af &  ar &  bg &  de &  el &  en &  es &  et &  eu &  fa &  fi &  fr &  he &  hi &  hu &  id &  it \\ \midrule
\our{} & 88.5 & 69.1 & 88.8 & 88.8 & 85.8 & 95.9 & 88.5 & 84.9 & 68.3 & 70.9 & 84.8 & 88.1 & 79.6 & 71.6 & 83.3 & 72.3 & 89.4 \\
\bottomrule
\end{tabular}
}
\renewcommand\tabcolsep{4.0pt}
\scalebox{0.9}{
\begin{tabular}{lccccccccccccccccc}
\toprule
 Model & ja &  kk &  ko &  mr &  nl &  pt &  ru &  ta &  te &  th &  tl &  tr &  ur &  vi &  yo &  zh & Avg \\ \midrule
\our{} & 51.1 & 75.3 & 53.8 & 80.3 & 89.3 & 87.6 & 88.9 & 62.3 & 85.9 & 60.2 & 90.1 & 74.8 & 63.3 & 55.9 & 24.2 & 67.9 & 76.0 \\
\bottomrule
\end{tabular}
}
\caption{Results on part-of-speech tagging.}
\label{table:udpos}
\end{table*}

\begin{table*}[b]
\centering
\small
\renewcommand\tabcolsep{4.0pt}
\scalebox{0.75}{
\begin{tabular}{lcccccccccccccccccccc}
\toprule
 Model & ar &   he &   vi &   id &   jv &   ms &   tl &   eu &   ml &   ta &   te &   af &   nl &   en &   de &   el &   bn &   hi &   mr &   ur \\ \midrule
\our{} & 57.7 & 54.3 & 72.5 & 49.7 & 56.9 & 68.3 & 72.0 & 53.1 & 68.6 & 58.0 & 54.6 & 76.3 & 82.1 & 84.2 & 77.9 & 76.4 & 73.1 & 69.2 & 64.9 & 65.8 \\
\bottomrule
\end{tabular}
}
\renewcommand\tabcolsep{4.0pt}
\scalebox{0.75}{
\begin{tabular}{lccccccccccccccccccccc}
\toprule
 Model & fa &   fr &   it &   pt &   es &   bg &   ru &   ja &   ka &   ko &   th &   sw &   yo &   my &   zh &   kk &   tr &   et &   fi &   hu &  Avg \\ \midrule
\our{} & 53.2 & 79.0 & 79.4 & 78.8 & 73.8 & 78.9 & 66.2 & 23.0 & 70.6 & 56.6 & 2.2 & 69.3 & 43.8 & 56.5 & 28.3 & 49.2 & 77.5 & 73.3 & 77.0 & 77.0 & 63.7 \\
\bottomrule
\end{tabular}
}
\caption{Results on WikiAnn named entity recognition.}
\label{table:wikiann}
\end{table*}

\begin{table*}[b]
\centering
\small
\renewcommand\tabcolsep{3.0pt}
\scalebox{0.75}{
\begin{tabular}{lcccccccccccc}
\toprule
 Model & en &  es &  de &  el &  ru &  tr &  ar &  vi &  th &  zh &  hi & Avg \\ \midrule
\our{} & 85.7 / 74.6 & 70.3 / 52.5 & 76.6 / 60.3 & 75.5 / 56.8 & 79.4 / 60.8 & 71.8 / 54.7 & 75.4 / 59.4 & 72.1 / 61.0 & 70.9 / 55.5 & 76.7 / 56.9 & 67.3 / 56.8 & 74.7 / 59.0 \\
\bottomrule
\end{tabular}
}
\caption{Results on XQuAD question answering.}
\label{table:xquad}
\end{table*}

\begin{table*}[b]
\centering
\small
\scalebox{0.9}{
\begin{tabular}{lcccccccc}
\toprule
 Model & en &  es &  de &  ar &  hi &  vi &  zh & Avg \\ \midrule
\our{} & 81.5 / 68.3 & 70.3 / 52.2 & 64.5 / 49.8 & 60.7 / 41.2 & 65.2 / 47.5 & 69.8 / 48.9 & 64.4 / 40.4 & 68.1 / 49.8 \\
\bottomrule
\end{tabular}
}
\caption{Results on MLQA question answering.}
\label{table:mlqa}
\end{table*}

\begin{table*}[b]
\centering
\small
\renewcommand\tabcolsep{5.0pt}
\scalebox{0.83}{
\begin{tabular}{lcccccccccc}
\toprule
 Model & en &  ar &  bn &  fi &  id &  ko &  ru &  sw &  te & Avg \\ \midrule
\our{} & 69.4 / 56.2 & 68.7 / 49.4 & 56.0 / 38.9 & 64.2 / 47.2 & 73.9 / 57.9 & 53.0 / 40.4 & 62.3 / 38.0 & 60.1 / 42.8 & 51.0 / 31.9 & 62.1 / 44.8 \\
\bottomrule
\end{tabular}
}
\caption{Results on TyDiQA question answering.}
\label{table:tydiqa}
\end{table*}

\begin{table*}[b]
\centering
\small
\scalebox{0.9}{
\begin{tabular}{lcccccccccccccccc}
\toprule
 Model & en &  fr &  es &  de &  el &  bg &  ru &  tr &  ar &  vi &  th &  zh &  hi &  sw &  ur & Avg \\ \midrule
\our{} & 86.7 & 80.6 & 81.0 & 78.8 & 77.4 & 78.8 & 77.4 & 75.2 & 73.9 & 76.9 & 73.8 & 77.0 & 71.9 & 67.1 & 66.6 & 76.2 \\
\bottomrule
\end{tabular}
}
\caption{Results on XNLI natural language inference.}
\label{table:xnli}
\end{table*}

\begin{table*}[b]
\centering
\small
\begin{tabular}{lcccccccc}
\toprule
 Model & en &  fr &  de &  es &  ja &  ko &  zh & Avg \\ \midrule
\our{} & 95.1 & 89.3 & 90.5 & 90.7 & 79.1 & 79.5 & 83.2 & 86.8 \\
\bottomrule
\end{tabular}
\caption{Results on PAWS-X cross-lingual paraphrase adversaries.}
\label{table:pawsx}
\end{table*}

\end{document}